\documentclass[10pt,a4paper]{article}

\usepackage{caption}
\usepackage{subcaption}
\usepackage{graphicx}
\graphicspath{{figures/}}
\usepackage{enumerate}

\usepackage{amssymb}
\usepackage{amsmath} 

\usepackage{array}
\usepackage{lipsum}
\usepackage{hyperref}
\usepackage{authblk}
\usepackage{setspace}
\usepackage{algorithm}
\usepackage{algpseudocode}
\usepackage{multirow}
\usepackage{booktabs}

\begin{document}
	
	\title{Benchmark for CEC 2024 Competition on Multiparty Multiobjective Optimization} 
	\author[1]{Wenjian Luo}
	\author[2]{Peilan Xu}
	\author[3]{Shengxiang Yang}
	\author[4]{Yuhui Shi}
	
	\affil[1]{School of Computer Science and Technology, Harbin Institute of Technology, Shenzhen 518055, China}
	\affil[2]{School of Artificial Intelligence, Nanjing University of Information Science and Technology, Nanjing 210044, China,}
	\affil[3]{School of Computer Science and Informatics, the De Montfort University, Leicester LE1 9BH, United Kingdom}
	\affil[4]{Department of Computer Science and Engineering, Southern University of Science and Technology, Shenzhen 518055, China}
	\affil[ ]{Email:
		\href{mailto:luowenjian@hit.edu.cn}{luowenjian@hit.edu.cn},
		\href{mailto:xpl@nuist.edu.cn}{xpl@nuist.edu.cn},
		\href{mailto:syang@dmu.ac.uk}{syang@dmu.ac.uk},
		\href{mailto:x.yao@cs.bham.ac.uk}{shiyh@sustech.edu.cn}}
	\date{\today} 
	\maketitle 

	\begin{abstract}
		\normalsize
		The competition focuses on Multiparty Multiobjective Optimization Problems (MPMOPs), where multiple decision makers have conflicting objectives, as seen in applications like UAV path planning. Despite their importance, MPMOPs remain understudied in comparison to conventional multiobjective optimization. The competition aims to address this gap by encouraging researchers to explore tailored modeling approaches. The test suite comprises two parts: problems with common Pareto optimal solutions and Biparty Multiobjective UAV Path Planning (BPMO-UAVPP) problems with unknown solutions. Optimization algorithms for the first part are evaluated using Multiparty Inverted Generational Distance (MPIGD), and the second part is evaluated using Multiparty Hypervolume (MPHV) metrics. The average algorithm ranking across all problems serves as a performance benchmark.
		
	\end{abstract}
	
	
	\clearpage
	
	\setcounter{tocdepth}{2}
	\tableofcontents

	\clearpage

	\section{Introduction}
	\label{sec:intro}
	
    While numerous heuristic algorithms have been effectively applied to a wide array of multiobjective optimization problems (MOPs) \cite{luo2022finding, lin2023novel, luo2019g, 10056413, song2021kriging}, practical domains such as business, science, and socio-political decision-making present a multitude of challenges. Among these challenges, the MOPs which involve multiple decision-makers (DMs), have not received much attention. Such problems are often regarded as general MOPs, wherein the optimization objectives of various decision-makers are collected, and multiple decision-makers are considered as a single role. Thus  the work in \cite{liu2020evolutionary} introduced a class of problems named multiparty multiobjective optimization problems (MPMOPs), building upon the foundation of MOPs.

    For better understanding MPMOPs, we need to present the definition of a MOP. Without loss of generality, a minimization MOP can be expressed in the following manner \cite{hong2018scalable}.

    \begin{equation}\label{eqt:MOP}
        \begin{aligned}
            \min \ & F(\mathbf{x}) = (f_1(\mathbf{x}), \dots, f_k(\mathbf{x})), \\
            \text{s.t.} & \left\{
                \begin{aligned}
                    g_i(\mathbf{x}) \le 0,\ & i= 1, \dots, p, \\
                    h_j(\mathbf{x}) =   0,\ & j= 1, \dots, q.
                \end{aligned} \right.
        \end{aligned}
    \end{equation}
    where $\mathbf{x} = (x_1, \dots, x_n)$ denotes the decision vector, while $(f_1, f_2, \dots, f_k)$ represents the objective functions. Additionally, $g_i (1 \le i \le q)$ and $h_j (1 \le j \le q)$ denote the inequality and equality constraint functions of the MOP, respectively.

    To introduce the notion of optimality in multi-objective optimization, several key definitions are essential. Given two vectors $ \mathbf{x}, \mathbf{y} \in \mathbb{R}^n $, $ \mathbf{x} $ dominates $ \mathbf{y} $ (denoted by $ \mathbf{x} \prec \mathbf{y} $) when the following two conditions are satisfied:
   \begin{itemize}
   		\item  $\forall i  \in \{1, 2, \dots, k\}$, $ f_i(\mathbf{x}) \leq f_i(\mathbf{y}) $;
   		\item $\exists i  \in \{1, 2, \dots, k\}$,  $ f_i(\mathbf{x}) < f_i(\mathbf{y}) $.
   \end{itemize}
	
	A decision vector $ \mathbf{x} $ is nondominated, i.e., Pareto-optimal, if there does not exist another vector $\mathbf{x}' $ such that $ \mathbf{x}' \prec \mathbf{x} $. The Pareto optimal Set $ PS $ comprises all Pareto-optimal vectors, denoted as $ PS = \{ \mathbf{x} \,|\, \mathbf{x} \text{ is Pareto-optimal}\} $. Similarly, the Pareto Front $ PF $ consists of the corresponding objective function values for the vectors in $ PS $, defined as $ PF = \{ f(\mathbf{x}) \in \mathbb{R}^k \,|\, \mathbf{x} \in P \} $.

    Then, the definition of a MPMOP is as follows \cite{liu2020evolutionary}:
    \begin{equation}\label{eqt:MPMOP}
      \min \ E(\mathbf{x}) = [F_1(\mathbf{x}), \dots, F_M(\mathbf{x})].
    \end{equation}
    where $F_i(\mathbf{x}),\ i=1, \dots, M$, represents a MOP defined as formula \eqref{eqt:MOP}.

    In the pursuit of optimal solutions for a MPMOP, the aim is to ensure that the solutions reach the $PF$ of all parties to the greatest extent possible. Consider a solution $\mathbf{x}^*$ deemed optimal for an MPMOP. For $\mathbf{x}^*$ to meet the criteria of being the best solution, there must not exist any other solution $ \mathbf{x} $ within the feasible region $ X $ that simultaneously satisfies the following two conditions: Firstly, for all parties involved, either $ \mathbf{x} \prec \mathbf{x}^* $ or they are nondominated by each other, indicating mutual nondominance. Secondly, for at least one party, $\mathbf{x} \prec \mathbf{x}^* $, signifying that $ \mathbf{x} $ is dominated by $ \mathbf{x}^* $ for that specific party's objectives.

    To efficiently address MPMOPs, several multiparty multiobjective evolutionary algorithms (MPMOEAs) have been developed \cite{liu2020evolutionary, she2021new, chang2022multiparty, song2022multiobjective, she2023privacy, chang2023biparty, she2023multiparty}.
    These algorithms employ variable strategies to enhance convergence towards different decision-makers using Pareto sorting methods.
    For instance, the pioneering OptMPNDS algorithm \cite{liu2020evolutionary} integrates the multiparty nondominated sorting (MPNDS) operator, which computes the multiparty rank by considering the ordinary Pareto sorting rank of distinct decision-makers. Furthermore, the OptMPNDS2 algorithm \cite{she2021new} incorporates the MPNDS2 operator, involving dual rounds of ordinary Pareto sorting.
    Chang \emph{et al.} \cite{chang2022multiparty} designed an MPMOEA based on MOEA/D, while Song \emph{et al.} \cite{song2022multiobjective} studied a class of multiparty multiobjective knapsack problems and proposed a new MPMOEA based on SPEA2.
    Privacy issues in MPMOPs have also been discussed by She et al. \cite{she2023privacy}. Additionally, studies by Chang et al. \cite{chang2023biparty} and She \emph{et al.} \cite{she2023multiparty} addressed the optimal power flow and distance minimization problems in real-world scenarios.
	
    In order to promote the development of MPMOEAs, a comprehensive benchmark suite is needed.
    In this report, we present the multiparty multiobjective benchmark suite for the competition, which comprises two parts.
    The first part, sourced from literature \cite{liu2020evolutionary}, consists of 11 problems featuring common Pareto optimal solutions. Specifically, these problems possess an overlapping set of Pareto optimal solutions among the parties, providing a basis for algorithm assessment.
    The second part encompasses six variations of biparty multiobjective UAV path planning (BPMO-UAVPP) problems, which is from \cite{chen2023MPUAV}. Optimal solutions for these BPMO-UAVPP scenarios remain unknown.
    It is noted that all functions are minimized.
    The website for the competition is available at \url{https://github.com/MiLab-HITSZ/CEC2024_MPMOP}.
	
	In the following contents, Section \ref{sec:multi} describes of the definition of two sets of multiparty multiobjective benchmark problems, and Section \ref{sec:ex} describes the evaluation criteria for the experiment.
	
	\section{Benchmark Suite}
	\label{sec:multi}
	
	This section describes the benchmark suites employed in the competition, which consist of two parts. The first part comprises 11 problems with common Pareto optimal solution. In other words, there exists a non-empty intersection of Pareto optimal solutions for these 11 problems, providing a basis for algorithm evaluation.
	The second part comprises six biparty multiobjective UAV path planning (BPMO-UAVPP) problems, for which the optimal solutions are currently unknown. Specifically, Section \ref{sec:BF} outlines the underlying multiobjective optimization functions utilized in the first benchmark suite, while Section \ref{sec:1stMPMOPs} introduces the first part of multiparty multiobjective benchmark suite consisting of 11 MPMOPs with common Pareto optimal solutions. Section \ref{sec:2ndMPMOPs} describes the second part of multiparty multiobjective benchmark suite, which includes six BPMO-UAVPP problems.

    \subsection{Basic Function} \label{sec:BF}
	These basic function is used to construct the 11 MPMO problems with common  Pareto optimal solutions \cite{liu2020evolutionary}, which are based on the CEC'2018 Competition on Dynamic Multiobjective Optimization \cite{jiang2018benchmark}.

    \subsubsection{BF1}
    \begin{equation}\label{eqt:g11}
      \left\{
        \begin{aligned}
        f_{11}(\mathbf{x}, t) & = d_1(\mathbf{x}, t)\left( \frac{1+t}{x_1}\right) \\
        f_{12}(\mathbf{x}, t) & = d_1(\mathbf{x}, t)\left( \frac{x_1}{1+t}\right) \\
        \end{aligned}
      \right.
    \end{equation}

    \begin{itemize}
      \item $d_1(\mathbf{x}, t) = 1 + \sum_{i=2}^{n}\left(x_i - \frac{1} {1 + \displaystyle e^{\alpha_t(x_1-2.5)}}\right)^2, $
      \item $\alpha_t = 5\cos(0.5\pi t)$,
      \item $x_1 \in [1, 4]$, $x_i \in [0, 1], i=2, \dots, n$.
      \item Pareto optimal set: $1 \le x_1 \le 4$, $x_i = 1 / (1 + e^{\alpha_t(x_1-2.5)}), i=2,\dots,n$
    \end{itemize}

    \subsubsection{BF2}
    \begin{equation}\label{eqt:g21}
      \left\{
        \begin{aligned}
        f_{21}(\mathbf{x}, t) & = d_2(\mathbf{x}, t)\left( x_1 + 0.1\sin(3\pi x_1)\right) \\
        f_{22}(\mathbf{x}, t) & = d_2(\mathbf{x}, t)\left( 1 - x_1 + 0.1\sin(3\pi x_1)\right)^{\alpha_t} \\
        \end{aligned}
      \right.
    \end{equation}

    \begin{itemize}
      \item $d_2(\mathbf{x}, t) = 1 + \sum_{i=2}^{n}\left(x_i -  \displaystyle \frac{\gamma_t \sin(4\pi x_1^{\beta_t})}{1+ |\gamma_t|}\right)^2,$
      \item $\alpha_t = 2.25 + 2\cos(2\pi t)$,
      \item $\beta_t = 1$,
      \item $\gamma_t = \sin(0.5\pi t)$,
      \item $x_1 \in [0, 1]$, $x_i \in [-1, 1], i=2, \dots, n$.
      \item Pareto optimal set: $0 \le x_1 \le 1$, $x_i = \gamma_t \sin(4\pi x_1^{\beta_t})/(1+|\gamma_t|), i=2,\dots,n$
    \end{itemize}

    \subsubsection{BF3}
    \begin{equation}\label{eqt:g3}
      \left\{
        \begin{aligned}
        f_{31}(\mathbf{x}, t) & = d_3(\mathbf{x}, t)\left( x_1 + \alpha_t \right) \\
        f_{32}(\mathbf{x}, t) & = d_3(\mathbf{x}, t)\left( 1 - x_1 + \alpha_t \right) \\
        \end{aligned}
      \right.
    \end{equation}

    \begin{itemize}
      \item $d_3(\mathbf{x}, t) = 1 + \sum_{i=2}^{n}\left(x_i - \cos(4t + x_1 + x_{i-1})\right)^2,$
      \item $\alpha_t = \max{0, (1/2\beta_t + 0.1)\sin(2\beta_t\pi x_1)}$,
      \item $\beta_t = 1 + \lfloor 10|\sin(0.5\pi t)| \rfloor$,

      \item $x_1 \in [0, 1]$, $x_i \in [-1, 1], i=2, \dots, n$.
      \item Pareto optimal set: $x_1 \in \cup_{i=1}^{\beta_t}[(2i-1)/2\beta_t, i/\beta_t]\cup\{0\}$, $x_i = \cos(4t+x_1+x_{i-1}), i=2,\dots,n$
    \end{itemize}

    \subsubsection{BF4}
    \begin{equation}\label{eqt:g3}
      \left\{
        \begin{aligned}
        f_{41}(\mathbf{x}, t) & = d_4(\mathbf{x}, t)[\sin(0.5\pi x_1)]^{\alpha_t} \\
        f_{42}(\mathbf{x}, t) & = d_4(\mathbf{x}, t)[\sin(0.5\pi x_2)\cos(0.5\pi x_1)]^{\alpha_t} \\
        f_{43}(\mathbf{x}, t) & = d_4(\mathbf{x}, t)[\cos(0.5\pi x_2)\cos(0.5\pi x_1)]^{\alpha_t}
        \end{aligned}
      \right.
    \end{equation}

    \begin{itemize}
      \item $d_4(\mathbf{x}, t) = 1 + \sum_{i=3}^{n}\left(x_i -  \displaystyle \frac{\sin(2\pi(x_1 + x_2))}{1+ |\beta_t|}\right)^2,$
      \item $\alpha_t = 2.25+2 \cos(0.5\pi t)$,
      \item $\beta_t = \sin(0.5 \pi t)$,

      \item $x_1, x_2 \in [0, 1]$, $x_i \in [-1, 1], i=3, \dots, n$.
      \item Pareto optimal set: $0 \le x_1, x_2 \le 1$, $x_i = \sin(2\pi(x_1+x_2))/(1+|\beta_t|), i=3,\dots,n$
    \end{itemize}

    \subsubsection{BF5}
    \begin{equation}\label{eqt:g3}
      \left\{
        \begin{aligned}
        f_{51}(\mathbf{x}, t) & = d_5(\mathbf{x}, t)\sin(y_1) \\
        f_{52}(\mathbf{x}, t) & = d_5(\mathbf{x}, t)\sin(y_2)\cos(y_1) \\
        f_{53}(\mathbf{x}, t) & = d_5(\mathbf{x}, t)\cos(y_2)\cos(y_1)
        \end{aligned}
      \right.
    \end{equation}

    \begin{itemize}
      \item $d_5(\mathbf{x}, t) = 1 + \sum_{i=3}^{n}\left(x_i - 0.5 \alpha_t x_1 \right)^2,$
      \item $\alpha_t = |\sin(0.5\pi t)|$,
      \item $y_i = (\pi/6)\alpha_t + (\pi/2 - (\pi/3)\alpha_t)x_i, i = 1, 2$,

      \item $x_i \in [0, 1], i=1, \dots, n$.
      \item Pareto optimal set: $0 \le x_1, x_2 \le 1$, $x_i = 0.5\alpha_t x_1, i=3,\dots,n$
    \end{itemize}

    \subsubsection{BF6}
    \begin{equation}\label{eqt:g3}
      \left\{
        \begin{aligned}
        f_{61}(\mathbf{x}, t) & = d_6(\mathbf{x}, t)\cos(0.5\pi x_1)\cos(0.5\pi x_2) \\
        f_{62}(\mathbf{x}, t) & = d_6(\mathbf{x}, t)\cos(0.5\pi x_1)\sin(0.5\pi x_2) \\
        f_{63}(\mathbf{x}, t) & = d_6(\mathbf{x}, t)\sin(0.5\pi x_1)
        \end{aligned}
      \right.
    \end{equation}

    \begin{itemize}
      \item $d_6(\mathbf{x}, t) = 1 + \sum_{i=3}^{n}\left(x_i - \sin(tx_1) \right)^2 + \arrowvert \prod_{j=1}^{2}\sin(\lfloor \alpha_t(2x_j - r) \rfloor )(\pi/2) \arrowvert, $
      \item $\alpha_t = \lfloor 10\sin(\pi t)\rfloor$,
      \item $r = 1 - \text{mod}(\alpha_t, 2)$,

      \item $x_1, x_2 \in [0, 1]$, $x_i \in [-1, 1], i=3, \dots, n$.
      \item Pareto optimal set: $\{x_1, x_2 \in [0, 1]\, |\, \prod_{j=1}^{2}\text{mod}(|\lfloor \alpha_t(2x_j-r)\rfloor |, 2) = 0\}$, $x_i = \sin(tx_1), i=3,\dots,n$
    \end{itemize}

	\subsection{MPMOPs with Common Pareto Optimal Solutions} \label{sec:1stMPMOPs}
	As mentioned in last subsection, these 11 MPMOP with common Pareto optimal solution are from Ref. \cite{liu2020evolutionary}, which are based on the CEC'2018 Competition on Dynamic Multiobjective Optimization \cite{jiang2018benchmark}.
	
	\subsubsection{MPMOP $E_1$}
	
    \begin{equation}\label{eqt:E1}
        E_1(\mathbf{x}) = [F_{11}(\mathbf{x}), F_{12}(\mathbf{x})]
    \end{equation}
    \[
    \left\{
      \begin{aligned}
        F_{11}(\mathbf{x}) & = [f_{11}(\mathbf{x}, 1), f_{12}(\mathbf{x}, 1)] \\
        F_{12}(\mathbf{x}) & = [f_{11}(\mathbf{x}, 2), f_{12}(\mathbf{x}, 2)]
      \end{aligned}
      \right.
    \]

	\subsubsection{MPMOP $E_2$}
	
    \begin{equation}\label{eqt:E1}
      E_2(\mathbf{x}) = [F_{21}(\mathbf{x}), F_{22}(\mathbf{x})]
    \end{equation}
    \[
    \left\{
      \begin{aligned}
        F_{21}(\mathbf{x}) & = [f_{21}(\mathbf{x}, 0), f_{22}(\mathbf{x}, 0)] \\
        F_{22}(\mathbf{x}) & = [f_{21}(\mathbf{x}, 3), f_{22}(\mathbf{x}, 3)]
      \end{aligned}
    \right.
    \]
	
	\subsubsection{MPMOP $E_3$}
	
    \begin{equation}\label{eqt:E1}
      E_3(\mathbf{x}) = [F_{31}(\mathbf{x}), F_{32}(\mathbf{x})]
    \end{equation}
    \[
    \left\{
      \begin{aligned}
        F_{31}(\mathbf{x}) & = [f_{31}(\mathbf{x}, 0), f_{32}(\mathbf{x}, 0)] \\
        F_{32}(\mathbf{x}) & = [f_{31}(\mathbf{x}, \pi/2), f_{32}(\mathbf{x}, \pi/2)]
      \end{aligned}
    \right.
    \]
	
	\subsubsection{MPMOP $E_4$}
	
    \begin{equation}\label{eqt:E1}
      E_4(\mathbf{x}) = [F_{41}(\mathbf{x}), F_{42}(\mathbf{x})]
    \end{equation}
    \[
    \left\{
      \begin{aligned}
        F_{41}(\mathbf{x}) & = [f_{41}(\mathbf{x}, 0), f_{42}(\mathbf{x}, 0), f_{43}(\mathbf{x}, 0)] \\
        F_{42}(\mathbf{x}) & = [f_{41}(\mathbf{x}, 1), f_{42}(\mathbf{x}, 1), f_{43}(\mathbf{x}, 1)]
      \end{aligned}
    \right.
    \]

    \subsubsection{MPMOP $E_5$}
	
    \begin{equation}\label{eqt:E1}
      E_5(\mathbf{x}) = [F_{51}(\mathbf{x}), F_{52}(\mathbf{x})]
    \end{equation}
    \[
    \left\{
      \begin{aligned}
        F_{51}(\mathbf{x}) & = [f_{51}(\mathbf{x}, 0), f_{52}(\mathbf{x}, 0), f_{53}(\mathbf{x}, 0)] \\
        F_{52}(\mathbf{x}) & = [f_{51}(\mathbf{x}, 1.5), f_{52}(\mathbf{x}, 1.5), f_{53}(\mathbf{x}, 1.5)]
      \end{aligned}
    \right.
    \]

    \subsubsection{MPMOP $E_6$}
	
    \begin{equation}\label{eqt:E1}
      E_6(\mathbf{x}) = [F_{61}(\mathbf{x}), F_{62}(\mathbf{x})]
    \end{equation}
    \[
    \left\{
      \begin{aligned}
        F_{61}(\mathbf{x}) & = [f_{61}(\mathbf{x}, 0), f_{62}(\mathbf{x}, 0), f_{63}(\mathbf{x}, 0)] \\
        F_{62}(\mathbf{x}) & = [f_{61}(\mathbf{x}, 1), f_{62}(\mathbf{x}, 1), f_{63}(\mathbf{x}, 1)]
      \end{aligned}
    \right.
    \]

	\subsubsection{MPMOP $E_7$}
	
    \begin{equation}\label{eqt:E1}
        E_7(\mathbf{x}) = [F_{71}(\mathbf{x}), F_{72}(\mathbf{x}), F_{73}(\mathbf{x})]
    \end{equation}
    \[
    \left\{
      \begin{aligned}
        F_{71}(\mathbf{x}) & = [f_{11}(\mathbf{x}, 0), f_{12}(\mathbf{x}, 0)] \\
        F_{72}(\mathbf{x}) & = [f_{11}(\mathbf{x}, 1), f_{12}(\mathbf{x}, 1)] \\
        F_{73}(\mathbf{x}) & = [f_{11}(\mathbf{x}, 2), f_{12}(\mathbf{x}, 2)]
      \end{aligned}
      \right.
    \]

	\subsubsection{MPMOP $E_8$}
	
    \begin{equation}\label{eqt:E1}
      E_8(\mathbf{x}) = [F_{81}(\mathbf{x}), F_{82}(\mathbf{x}), F_{83}(\mathbf{x})]
    \end{equation}
    \[
    \left\{
      \begin{aligned}
        F_{81}(\mathbf{x}) & = [f_{21}(\mathbf{x}, 0), f_{22}(\mathbf{x}, 0)] \\
        F_{81}(\mathbf{x}) & = [f_{21}(\mathbf{x}, 1), f_{22}(\mathbf{x}, 1)] \\
        F_{83}(\mathbf{x}) & = [f_{21}(\mathbf{x}, 3), f_{22}(\mathbf{x}, 3)]
      \end{aligned}
    \right.
    \]
	
	\subsubsection{MPMOP $E_9$}
	
    \begin{equation}\label{eqt:E1}
      E_9(\mathbf{x}) = [F_{91}(\mathbf{x}), F_{92}(\mathbf{x}), F_{93}(\mathbf{x})]
    \end{equation}
    \[
    \left\{
      \begin{aligned}
        F_{91}(\mathbf{x}) & = [f_{41}(\mathbf{x}, 0), f_{42}(\mathbf{x}, 0), f_{43}(\mathbf{x}, 0)] \\
        F_{92}(\mathbf{x}) & = [f_{41}(\mathbf{x}, 0.5), f_{42}(\mathbf{x}, 0.5), f_{43}(\mathbf{x}, 0.5)] \\
        F_{93}(\mathbf{x}) & = [f_{41}(\mathbf{x}, 1), f_{42}(\mathbf{x}, 1), f_{43}(\mathbf{x}, 1)]
      \end{aligned}
    \right.
    \]

    \subsubsection{MPMOP $E_{10}$}
	
    \begin{equation}\label{eqt:E1}
      E_{10}(\mathbf{x}) = [F_{10, 1}(\mathbf{x}), F_{10, 2}(\mathbf{x}), F_{10, 3}(\mathbf{x})]
    \end{equation}
    \[
    \left\{
      \begin{aligned}
        F_{10,1}(\mathbf{x}) & = [f_{51}(\mathbf{x}, 0), f_{52}(\mathbf{x}, 0), f_{53}(\mathbf{x}, 0)] \\
        F_{10,1}(\mathbf{x}) & = [f_{51}(\mathbf{x}, 1), f_{52}(\mathbf{x}, 1), f_{53}(\mathbf{x}, 1)] \\
        F_{10,2}(\mathbf{x}) & = [f_{51}(\mathbf{x}, 1.5), f_{52}(\mathbf{x}, 1.5), f_{53}(\mathbf{x}, 1.5)]
      \end{aligned}
    \right.
    \]

    \subsubsection{MPMOP $E_{11}$}
	
    \begin{equation}\label{eqt:E1}
      E_{11}(\mathbf{x}) = [F_{11,1}(\mathbf{x}), F_{11,2}(\mathbf{x})]
    \end{equation}
    \[
    \left\{
      \begin{aligned}
        F_{11,1}(\mathbf{x}) & = [f_{61}(\mathbf{x}, 0), f_{62}(\mathbf{x}, 0), f_{63}(\mathbf{x}, 0)] \\
        F_{11,2}(\mathbf{x}) & = [f_{61}(\mathbf{x}, 1), f_{62}(\mathbf{x}, 1), f_{63}(\mathbf{x}, 1)] \\
        F_{11,3}(\mathbf{x}) & = [f_{61}(\mathbf{x}, 1.5), f_{62}(\mathbf{x}, 1.5), f_{63}(\mathbf{x}, 1.5)] \\
      \end{aligned}
    \right.
    \]
	
	\subsection{Biparty Multi-objective UAV Path Planning} \label{sec:2ndMPMOPs}

    This report introduces a biparty multi-objective UAV path planning (BP-UAVPP) model featuring efficiency and safety decision makers, which is from \cite{chen2023MPUAV}. The efficiency component emphasizes metrics like path length, height, flight energy consumption, and mission hover point distance. Conversely, the safety aspect targets third-party risk factors such as fatality risk, property risk, and noise pollution.

    \subsubsection{Efficiency Decision-Maker}\label{eff}

For the efficiency decision-maker, maximizing UAV flight efficiency is paramount. This objective entails minimizing various factors, including UAV path length, height changes, flight energy consumption, and the total sum of mission hover point distances. Each discrete trajectory point along the path, denoted as $\vec{p_i}=(x_i,y_i,z_i)$ for the $i$th point, contributes to defining track segments, expressed as $g_i=\vec{p}_{i+1}-\vec{p}_{i}$.

\textbf{Path length \cite{pengDecompositionbasedConstrainedMultiobjective2022}}:
The first objective during the UAV mission is to minimize the path length of all segments traversed by the UAV from the starting point to the target point:
\begin{equation}
	f_{\textup{length}} = \sum_{i=0}^{n-1}||g_i||.
\end{equation}

\textbf{Flight energy consumption \cite{ghambariEnhancedNSGAIIMultiobjective2020}}:
The second objective during the UAV mission is to minimize the flight energy consumption of all segments traversed by the UAV from the starting point to the target point:

\begin{equation}
	f_{\textup{fuel}} = \sum_{i=0}^{n-1} {fuel}_i,
\end{equation}
where ${fuel}_i$ represents the flight energy consumption of the $i$-th track segment. This consumption depends on the flight state, hovering power usage, and the energy needed to overcome gravity. Specifically, ${fuel}_i$ is defined as:

\begin{equation}
	fuel_i = W^{\frac{3}{2}}\sqrt{\frac{G^3}{2\rho_i \zeta n}}\frac{||g_i||}{V}+\Delta^{+}WG,
\end{equation}
where $W$ represents the weight of the UAV and battery, $\rho$ denotes the fluid density of air, $G$ stands for gravity, $\Delta^{+}$ signifies the altitude increase, $n$ indicates the number of rotors, $||g_i||$ represents the track length, $V$ denotes the flight speed, and $\zeta $ refers to the area of the rotating blade disk. The atmospheric density $\rho_i$ as a function of altitude is expressed as $\rho_i=\rho_0\exp(-(z_{i+1}+z_{i})/(2*10.7)),$ where $\rho_0$ is the atmospheric density at sea level and is considered as 1.225$kg/m^3$ in this report.

\textbf{Path height changes \cite{aitsaadiUAVPathPlanning2022a}}:
The third objective of the UAV mission was to minimize the height change at each trajectory point, thereby reducing the need for UAV ascent and descent, which extends the lifespan of the UAV power system:
\begin{equation}
	\label{eqt:height_changes}
	f_{\textup{height}} = \sum_{i=0}^{n-1} |z_{i+1}-z_{i}| .
\end{equation}

\textbf{Mission hover point distance \cite{shen2022energy}}:
During the UAV mission, the fourth objective aimed to minimize the sum of distances to all preset UAV hover points (UHPs) while hovering, thereby enhancing the UAV's ability to collect sensor data.

\begin{equation}
	f_{\textup{distance}} = \sum_{k=0}^{K-1} \mathop{\min}\limits_{i} ||\vec{p_i}-p_k^{job}||,
\end{equation}
where $p_k^{job}$ is the $k$-th preset UHP and $\vec{p_i}$ is the $i$th discrete trajectory point.

\subsubsection{Safety Decision-Maker}\label{safef}
The primary concern for the safety decision-maker is to minimize the risk posed to urban residents and property by UAV operations. As outlined in Ref. \cite{pangUAVPathOptimization2022a}, the key objectives of safety decision-making include minimizing risks to pedestrians, vehicles, and property, as well as mitigating noise pollution.

\textbf{Fatality Risks \cite{pangUAVPathOptimization2022a}}:
The first objective of the UAV safety decision maker is to minimize fatality risk costs, encompassing risks to pedestrians as well as risks to vehicles.
\begin{equation}
	f_{\textup{fatal}} = \sum_{i=0}^{n}c_{r_p}(x_i,y_i,z_i)+\sum_{i=0}^{n}c_{r_v}(x_i,y_i,z_i),
\end{equation}
where $c_{r_p}$ and $c_{r_v}$ represent the fatality risk cost associated with pedestrian and vehicles fatalities, which are expressed as follows:

\begin{equation}
  \left \{
    \begin{aligned}
        c_{r_p}(x,y,z) = & P_{crash}S_{h}\sigma_{p}(x,y)R^{P}_{f}(z)	\\
        c_{r_v}(x,y,z) = & P_{crash}S_{h}\sigma_{v}(x,y)R^{V}_{f}(z)
    \end{aligned}
  \right.
\end{equation}

In the above formula, the crash probability, $P_{\text{crash}}$, primarily hinges on the UAV's reliability, encompassing both hardware and software reliability. $S_{h}$ denotes the UAV's crash impact area size, while $\sigma_{p}(x,y)$ signifies the population density within the administrative unit, constituting a location-dependent factor. $R^{P}_{f}(z)$ correlates with the kinetic energy of the impact and the obscuration factor, whereas $R^{V}_{f}(z)$ represents the vehicle's impact energy parameters.

\textbf{Property Risk \cite{pangUAVPathOptimization2022a}}
The second objective of the UAV safety decision maker is to minimize property risk:
\begin{equation}
	f_{\textup{eco}} = \sum_{i=0}^{n}c_{r_p,d}(z_i),
\end{equation}
where $c_{r_p,d}$ represents the property risk index, defined as follows.
\begin{equation}
	c_{r_p,d}(z_i)=\begin{cases}
		\psi(e^\mu) &{\text{if}}\ 0< z_i \le e^\mu \\
		\psi(z_i) &{\text{otherwise.}}
	\end{cases},
\end{equation}
\begin{equation}
	\psi(z_i;\mu,\sigma)=\frac{1}{z_i \sigma\sqrt{2\pi}}e^{-\frac{(\ln z_i-\mu)^2}{2\sigma^2}},
\end{equation}
where the function $\psi$ is a distribution function of the flight height. $\mu$ and $\sigma$ are the lognormal distribution parameters of the building height.

\textbf{Noise Pollution \cite{pangUAVPathOptimization2022a}}:
The third objective of the UAV safety decision maker is to minimize noise pollution, which could be expressed as the approximate value of spherical propagation, as follows.
\begin{equation}
	f_{\textup{noise}} = \sum_{i=0}^{n} C_{noise}(x_i,y_i,z_i).
\end{equation}

In the above formula, $C_{noise}$ is the noise impact cost of UAV operations in a given airspace cell, and

\begin{equation} C_{noise}(x_i,y_i,z_i)=k\sigma_{p}(x_i,y_i)L_h\frac{1}{(z_i)^2+d^2},
\end{equation}
where $k$ is the conversion factor from the sound intensity to the sound level, $L_h$ is the noise produced by the UAV, and $\sigma_{p}(x_i,y_i)$ is the density of people at track point $(x_i,y_i,z_i)$. $d$ is the distance between the UAV and the area of interest; if the flight altitude exceeds a certain threshold, the noise is small, and its impact will not be included in the calculation of pollution. The noise threshold was set to 40 dB in the experiment \cite{torijaEffectsHoveringUnmanned2020}.

\subsubsection{Constraints}\label{con}
The following constraints need to be satisfied during a UAV mission.

\textbf{Terrain Constraints}: The UAV can only fly within a certain altitude range,
\begin{equation}
	H_{min} \leq z_i \leq H_{max}.
\end{equation}

\textbf{Performance Constraints \cite{aitsaadiUAVPathPlanning2022a}}: Given the UAV's power limitations, the turning angle between segments must not exceed the maximum turning angle, and the slope angle between segments must not surpass the maximum slope angle. These constraints are expressed as follows.
\begin{equation}
  \left\{
    \begin{aligned}
	   |\alpha_i| & \leq \alpha_{\max}, \\
        |\beta_i| & \leq \beta_{\max},
    \end{aligned}
\right.
\end{equation}

\begin{equation}
  \left\{
    \begin{aligned}
	   \alpha_i = & \arccos{\frac{g_i'*g_{i+1}'}{||g_i'||*||g_{i+1}'||}}, \\
        \beta_i = & \arctan{\frac{z_{i+1}-z_i}{||g_i'||}},
    \end{aligned}
\right.
\end{equation}
where $\alpha_i$ and $\beta_i$ denote the turning angle and slope angle between the $i$-th and $(i+1)$-th segments, respectively. $\alpha_{max}$ and $\beta_{max}$ represent the maximum turning and slope angles of the UAV, while $g_i'$ indicates the projection of the $i$-th track segment in the plane.

\subsubsection{Test Case}
The second part of benchmark suite comprises six BP-UAVPP scenarios. Detailed information on these benchmark cases is presented in Table \ref{table3}.

In generating building data, lognormal distribution characterizes building heights, with parameters $\mu = 3.04670$ and $\sigma = 0.76023$ \cite{pangUAVPathOptimization2022a}. Population distribution centers around the core metropolitan area. Following Pang \emph{ et al.} \cite{pangUAVPathOptimization2022a}, a radial basis model simulates population density for data generation.

UAV fuel consumption parameters align with those in Ref. \cite{ghambariEnhancedNSGAIIMultiobjective2020}, while third-party risk parameters match those in Ref. \cite{pangUAVPathOptimization2022a}. Additional parameters include: maximum turning angle $\alpha_{max} = \pi/3$, maximum slope angle $\beta_{max} = \pi/4$, atmospheric density $\rho_{A} = 1.225 kg/m^3$, flight speed $v = 10 m/s $, rotating slope area $S_b = 0.1 m^2$, number of rotating paddles $n = 4$, and UAV weight $m = 1.38 kg$ \cite{pangUAVPathOptimization2022a}. The mission begins at (1,1) and ends at (45, 45), with preset UHPs at (25, 30), (34, 20), and (40, 35).

    \begin{table}[H]
	\centering
	\caption{Details of the Test Case}
	\begin{tabular}{ccc}
		\toprule
		Problems &  Efficiency DM Objectives   & Safety DM Objectives        \\
		\cline{1-3}
		$C1$   &  $F_{\text{eff}}=(f_{\textup{length}},f_{\textup{distance}})$  & $F_{\text{safe}}=(f_{\textup{fatal}},f_{\textup{eco}})$   	\\
		$C2$   &  $F_{\text{eff}}=(f_{\textup{length}}+f_{\textup{height}},f_{\textup{distance}})$  & $F_{\text{safe}}=(f_{\textup{fatal}},f_{\textup{eco}})$      \\
		$C3$  &   $F_{\text{eff}}=(f_{\textup{fuel}},f_{\textup{distance}})$  & $F_{\text{safe}}=(f_{\textup{fatal}},f_{\textup{eco}})$     \\
		$C4$  &   $F_{\text{eff}}=(f_{\textup{length}},f_{\textup{distance}})$  & $F_{\text{safe}}=(f_{\textup{fatal}},f_{\textup{noise}})$      \\
		$C5$  &  $F_{\text{eff}}=(f_{\textup{length}}+f_{\textup{height}},f_{\textup{distance}})$ &
		$F_{\text{safe}}=(f_{\textup{fatal}},f_{\textup{noise}})$ \\
		$C6$  &  $F_{\text{eff}}=(f_{\textup{fuel}},f_{\textup{distance}})$  & $F_{\text{safe}}=(f_{\textup{fatal}},f_{\textup{noise}})$  \\
		\bottomrule
	\end{tabular}
	\label{table3}
    \end{table}
	
	\section{Experimental Criteria}
	\label{sec:ex}
	
	\subsection{General Settings}
	
	The settings for MPMOPs are listed as follows.
	
	\begin{enumerate}
        \item \textbf{Dimension}: The dimensions of the decision variables for the first set of benchmark suites include 10, 30, 50.
		\item \textbf{Runs}: 30. The random seeds of all runs should be fixed from 1 to 30.
		\item \textbf{Maximum Fitness evaluations}: the maximum fitness evaluations of the first set of benchmark suites is set to $1000\cdot D \cdot M$, and the second set of benchmark suites is set to 100000.
		\item \textbf{Termination condition}: all the fitness evaluations are consumed.
	\end{enumerate}
	
	\subsection{Performance Metric}
    For the first part of the benchmark, this report utilizes the the multiparty inverted generational distance (MPIGD \cite{liu2020evolutionary}) to measure the performance of the algorithms. MPIGD is defined as
    \begin{equation}
    	\text{MPIGD}(P^{MP},P)=\frac{\sum_{v \in P^{MP}} d(v,P)}{|P^{MP}|},
    \end{equation}
    \begin{equation}
    	d(v,P) = \min_{s \in P}(\sum_{j=1}^{M}\sqrt{(v_{j1}-s_{j1})^2+\dots+(v_{jm_j}-s_{jm_j})^2}),
    \end{equation}	
    where $P^{MP}$ represents the true PF of the MPMOP and $P$ is the PF obtained by the algorithms. $d(v, P)$ represents the minimum distance between $v$ from $P^{MP}$ and points from $P$. Respectively, $(v_{j1}, . . . , v_{jm})$ means the $m$ objectives of the $j$-th DM for solution $v$, and $(s_{j1}, . . . , s_{jm})$ means the same for solution $s$.

    In the second part of benchmark suite, there is often no definitive true PF available for calculating the MPIGD metric. To evaluate the performance of various algorithms on multi-party multi-objective optimization problems (MPMOPs), this report utilizes the multiparty hypervolume (MPHV) metric, derived from the hypervolume (HV) metric \cite{zitzler1999multiobjective}. For general MOPs, the HV metric quantifies the hypervolume formed by the normalized solution set. Let $HV_i$ denote the HV metric of the solution set regarding the objectives of the $i$-th decision maker. Then, MPHV is defined as the average of $HV_i$ across all decision makers, expressed as:
    \begin{equation}
        \text{MPHV} = \sum_{i=1}^K HV_i.
    \end{equation}
	
	\subsection{Results Format}
	Participants are required to submit the best, median, worst, mean, and standard deviation of the 30 runs of the MPIGD for the first set of benchmark problems, as well as the MPHV results for the second set of benchmark problems, in the provided table below.
	
\begin{table}[htbp]
  \centering
  \caption{Experimental Results}
    \begin{tabular}{|c|l|c|c|c|c|c|c|}
    \hline
          &       & $E_1$  & $E_2$  & $E_3$  & $E_4$  & $E_5$  & $E_6$ \\
    \hline
    \multirow{5}[10]{*}{d=10} & Best  &       &       &       &       &       &  \\
\cline{2-8}          & Median &       &       &       &       &       &  \\
\cline{2-8}          & Worst &       &       &       &       &       &  \\
\cline{2-8}          & Mean  &       &       &       &       &       &  \\
\cline{2-8}          & StDev &       &       &       &       &       &  \\
    \hline
    \multirow{5}[10]{*}{d=30} & Best  &       &       &       &       &       &  \\
\cline{2-8}          & Median &       &       &       &       &       &  \\
\cline{2-8}          & Worst &       &       &       &       &       &  \\
\cline{2-8}          & Mean  &       &       &       &       &       &  \\
\cline{2-8}          & StDev &       &       &       &       &       &  \\
    \hline
    \multirow{5}[10]{*}{d=50} & Best  &       &       &       &       &       &  \\
\cline{2-8}          & Median &       &       &       &       &       &  \\
\cline{2-8}          & Worst &       &       &       &       &       &  \\
\cline{2-8}          & Mean  &       &       &       &       &       &  \\
\cline{2-8}          & StDev &       &       &       &       &       &  \\
    \hline
    \multicolumn{1}{r}{} & \multicolumn{1}{r}{} & \multicolumn{1}{c}{} & \multicolumn{1}{c}{} & \multicolumn{1}{c}{} & \multicolumn{1}{c}{} & \multicolumn{1}{c}{} & \multicolumn{1}{c}{} \\
    \hline
          &       & $E_7$  & $E_8$  & $E_9$  & $E_{10}$ & $E_{11}$ & - \\
    \hline
    \multirow{5}[10]{*}{d=10} & Best  &       &       &       &       &       &  \\
\cline{2-8}          & Median &       &       &       &       &       &  \\
\cline{2-8}          & Worst &       &       &       &       &       &  \\
\cline{2-8}          & Mean  &       &       &       &       &       &  \\
\cline{2-8}          & StDev &       &       &       &       &       &  \\
    \hline
    \multirow{5}[10]{*}{d=30} & Best  &       &       &       &       &       &  \\
\cline{2-8}          & Median &       &       &       &       &       &  \\
\cline{2-8}          & Worst &       &       &       &       &       &  \\
\cline{2-8}          & Mean  &       &       &       &       &       &  \\
\cline{2-8}          & StDev &       &       &       &       &       &  \\
    \hline
    \multirow{5}[10]{*}{d=50} & Best  &       &       &       &       &       &  \\
\cline{2-8}          & Median &       &       &       &       &       &  \\
\cline{2-8}          & Worst &       &       &       &       &       &  \\
\cline{2-8}          & Mean  &       &       &       &       &       &  \\
\cline{2-8}          & StDev &       &       &       &       &       &  \\
    \hline

    \multicolumn{1}{r}{} & \multicolumn{1}{r}{} & \multicolumn{1}{r}{} & \multicolumn{1}{r}{} & \multicolumn{1}{r}{} & \multicolumn{1}{r}{} & \multicolumn{1}{r}{} & \multicolumn{1}{r}{} \\
    \hline
          &       & $C_1$  & $C_2$  & $C_3$  & $C_4$  & $C_5$  & $C_6$ \\
    \hline
    \multirow{5}[10]{*}{d=88} & Best  &       &       &       &       &       &  \\
\cline{2-8}          & Median &       &       &       &       &       &  \\
\cline{2-8}          & Worst &       &       &       &       &       &  \\
\cline{2-8}          & Mean  &       &       &       &       &       &  \\
\cline{2-8}          & StDev &       &       &       &       &       &  \\
    \hline

    \end{tabular}%
  \label{tab:addlabel}%
\end{table}%

	
	
	\bibliography{bibliography/sample}{}

\begin{thebibliography}{10}
\providecommand{\url}[1]{#1}
\csname url@samestyle\endcsname
\providecommand{\newblock}{\relax}
\providecommand{\bibinfo}[2]{#2}
\providecommand{\BIBentrySTDinterwordspacing}{\spaceskip=0pt\relax}
\providecommand{\BIBentryALTinterwordstretchfactor}{4}
\providecommand{\BIBentryALTinterwordspacing}{\spaceskip=\fontdimen2\font plus
\BIBentryALTinterwordstretchfactor\fontdimen3\font minus
  \fontdimen4\font\relax}
\providecommand{\BIBforeignlanguage}[2]{{%
\expandafter\ifx\csname l@#1\endcsname\relax
\typeout{** WARNING: IEEEtran.bst: No hyphenation pattern has been}%
\typeout{** loaded for the language `#1'. Using the pattern for}%
\typeout{** the default language instead.}%
\else
\language=\csname l@#1\endcsname
\fi
#2}}
\providecommand{\BIBdecl}{\relax}
\BIBdecl

\bibitem{luo2022finding}
W.~Luo, L.~Shi, X.~Lin, J.~Zhang, M.~Li, and X.~Yao, ``Finding top-{K}
  solutions for the decision-maker in multiobjective optimization,''
  \emph{Information Sciences}, vol. 613, pp. 204--227, 2022.

\bibitem{lin2023novel}
X.~Lin, W.~Luo, N.~Gu, and Q.~Zhang, ``A novel dynamic reference point model
  for preference-based evolutionary multiobjective optimization,''
  \emph{Complex \& Intelligent Systems}, vol.~9, no.~2, pp. 1415--1437, 2023.

\bibitem{luo2019g}
W.~Luo, L.~Shi, X.~Lin, and C.~A.~C. Coello, ``The $\hat{g}$-dominance relation
  for preference-based evolutionary multi-objective optimization,'' in
  \emph{2019 IEEE Congress on Evolutionary Computation (CEC)}.\hskip 1em plus
  0.5em minus 0.4em\relax IEEE, 2019, pp. 2418--2425.

\bibitem{10056413}
S.~Liu, Q.~Lin, J.~Li, and K.~C. Tan, ``A survey on learnable evolutionary
  algorithms for scalable multiobjective optimization,'' \emph{IEEE
  Transactions on Evolutionary Computation}, vol.~27, no.~6, pp. 1941--1961,
  2023.

\bibitem{song2021kriging}
Z.~Song, H.~Wang, C.~He, and Y.~Jin, ``A kriging-assisted two-archive
  evolutionary algorithm for expensive many-objective optimization,''
  \emph{IEEE Transactions on Evolutionary Computation}, vol.~25, no.~6, pp.
  1013--1027, 2021.

\bibitem{liu2020evolutionary}
W.~Liu, W.~Luo, X.~Lin, M.~Li, and S.~Yang, ``Evolutionary approach to
  multiparty multiobjective optimization problems with common pareto optimal
  solutions,'' in \emph{2020 IEEE Congress on Evolutionary Computation
  (CEC)}.\hskip 1em plus 0.5em minus 0.4em\relax IEEE, 2020, pp. 1--9.

\bibitem{hong2018scalable}
W.~Hong, K.~Tang, A.~Zhou, H.~Ishibuchi, and X.~Yao, ``A scalable
  indicator-based evolutionary algorithm for large-scale multiobjective
  optimization,'' \emph{IEEE Transactions on Evolutionary Computation},
  vol.~23, no.~3, pp. 525--537, 2018.

\bibitem{she2021new}
Z.~She, W.~Luo, Y.~Chang, X.~Lin, and Y.~Tan, ``A new evolutionary approach to
  multiparty multiobjective optimization,'' in \emph{International Conference
  on Swarm Intelligence}.\hskip 1em plus 0.5em minus 0.4em\relax Springer,
  2021, pp. 58--69.

\bibitem{chang2022multiparty}
Y.~Chang, W.~Luo, X.~Lin, Z.~She, and Y.~Shi, ``Multiparty multiobjective
  optimization by moea/d,'' in \emph{2022 IEEE Congress on Evolutionary
  Computation (CEC)}.\hskip 1em plus 0.5em minus 0.4em\relax IEEE, 2022, pp.
  01--08.

\bibitem{song2022multiobjective}
Z.~Song, W.~Luo, X.~Lin, Z.~She, and Q.~Zhang, ``On multiobjective knapsack
  problems with multiple decision makers,'' in \emph{2022 IEEE Symposium Series
  on Computational Intelligence (SSCI)}.\hskip 1em plus 0.5em minus 0.4em\relax
  IEEE, 2022, pp. 156--163.

\bibitem{she2023privacy}
Z.~She, W.~Luo, Y.~Chang, Z.~Song, and Y.~Shi, ``On the privacy issue of
  evolutionary biparty multiobjective optimization,'' in \emph{International
  Conference on Swarm Intelligence}.\hskip 1em plus 0.5em minus 0.4em\relax
  Springer, 2023, pp. 371--382.

\bibitem{chang2023biparty}
Y.~Chang, W.~Luo, X.~Lin, Z.~Song, and C.~A.~C. Coello, ``Biparty
  multiobjective optimal power flow: The problem definition and an evolutionary
  approach,'' \emph{Applied Soft Computing}, vol. 146, p. 110688, 2023.

\bibitem{she2023multiparty}
Z.~She, W.~Luo, X.~Lin, Y.~Chang, and Y.~Shi, ``Multiparty distance
  minimization: Problems and an evolutionary approach,'' \emph{Swarm and
  Evolutionary Computation}, vol.~83, p. 101415, 2023.

\bibitem{jiang2018benchmark}
S.~Jiang, S.~Yang, X.~Yao, K.~C. Tan, M.~Kaiser, and N.~Krasnogor, ``Benchmark
  functions for the cec'2018 competition on dynamic multiobjective
  optimization,'' Newcastle University, Tech. Rep., 2018.

\bibitem{chen2023MPUAV}
K.~Chen, W.~Luo, X.~Lin, Z.~Song, and C.~Yatong, ``Evolutionary biparty
  multiobjective {UAV} path planning: Problems and empirical comparisons,''
  \emph{IEEE Transactions on Emerging Topics in Computational Intelligence},
  2023 accepted).

\bibitem{pengDecompositionbasedConstrainedMultiobjective2022}
C.~Peng and S.~Qiu, ``A decomposition-based constrained multi-objective
  evolutionary algorithm with a local infeasibility utilization mechanism for
  {UAV} path planning,'' \emph{Applied Soft Computing}, vol. 118, p. 108495,
  2022.

\bibitem{ghambariEnhancedNSGAIIMultiobjective2020}
S.~Ghambari, M.~Golabi, J.~Lepagnot, M.~Brevilliers, L.~Jourdan, and
  L.~Idoumghar, ``An enhanced {NSGA-II} for multiobjective {UAV} path planning
  in urban environments,'' in \emph{Proceeding of 2020 IEEE 32nd International
  Conference on Tools with Artificial Intelligence (ICTAI)}.\hskip 1em plus
  0.5em minus 0.4em\relax IEEE, 2020, pp. 106--111.

\bibitem{aitsaadiUAVPathPlanning2022a}
A.~Ait~Saadi, A.~Soukane, Y.~Meraihi, A.~Benmessaoud~Gabis, S.~Mirjalili, and
  A.~{Ramdane-Cherif}, ``{UAV} path planning using optimization approaches: A
  survey,'' \emph{Archives of Computational Methods in Engineering}, 2022.

\bibitem{shen2022energy}
L.~Shen, N.~Wang, D.~Zhang, J.~Chen, X.~Mu, and K.~M. Wong, ``Energy-aware
  dynamic trajectory planning for {UAV}-enabled data collection in {mMTC}
  networks,'' \emph{IEEE Transactions on Green Communications and Networking},
  vol.~6, no.~4, pp. 1957--1971, 2022.

\bibitem{pangUAVPathOptimization2022a}
B.~Pang, X.~Hu, W.~Dai, and K.~H. Low, ``{UAV} path optimization with an
  integrated cost assessment model considering third-party risks in
  metropolitan environments,'' \emph{Reliability Engineering \& System Safety},
  vol. 222, p. 108399, Jun. 2022.

\bibitem{torijaEffectsHoveringUnmanned2020}
A.~J. Torija, Z.~Li, and R.~H. Self, ``Effects of a hovering unmanned aerial
  vehicle on urban soundscapes perception,'' \emph{Transportation Research Part
  D: Transport and Environment}, vol.~78, p. 102195, Jan. 2020.

\bibitem{zitzler1999multiobjective}
E.~Zitzler and L.~Thiele, ``Multiobjective evolutionary algorithms: a
  comparative case study and the strength pareto approach,'' \emph{IEEE
  transactions on Evolutionary Computation}, vol.~3, no.~4, pp. 257--271, 1999.

\end{thebibliography}
	\bibliographystyle{IEEEtran}
	
\end{document}